\documentclass[runningheads]{llncs}

\usepackage{eccv}

\usepackage{eccvabbrv}

\usepackage{graphicx}
\usepackage{booktabs}

\usepackage[accsupp]{axessibility}  

\usepackage{hyperref}

\usepackage{orcidlink}

\usepackage{amsthm}

\theoremstyle{definition}

\usepackage{algorithm}
\usepackage{algpseudocode} 

\usepackage{enumitem}
\usepackage[table]{xcolor}
\usepackage{wrapfig}

\usepackage{microtype}
\microtypesetup{protrusion=true, expansion=false}

\begin{document}

\title{InfraNet: Quality-Aware RGB Guidance for Efficient Infrared Object Detection} 

\titlerunning{InfraNet for Infrared Object Detection}


\author{
Zichao Feng$^{1*}$ \and
Haodong Zhu$^{1,2*}$ \and
Jingying Yang$^{1}$ \and
Sheng Xu$^{3\dagger}$ \and\\
Yangyang Ren$^{1,2}$ \and
Yuguang Yang$^1$ \and
Xuhui Liu$^4$ \and
Juan Zhang$^1$ \and\\
Tian Wang$^1$ \and
Linlin Yang$^{3}$ \and
Baochang Zhang$^{1}$ \\
}

\authorrunning{Feng et al.}

\institute{Beihang University, Beijing, China \and
Zhongguancun Academy, Beijing, China\and
Communication University of China, Beijing, China\and
King Abdullah University of Science and Technology, Thuwal, Kingdom of Saudi Arabia
}

\maketitle
\let\thefootnote\relax
\footnotetext{$^*$ Equal contribution.~~~\{fengzichao, HaodongZhu\}@buaa.edu.cn}
\footnotetext{$^\dagger$ Corresponding author.~~~xsheng001@gmail.com}

\begin{abstract}
Robust object detection under adverse visual conditions remains a long-standing challenge for multi-modal perception systems.
{Existing fusion-based methods typically require both RGB and infrared (IR) inputs, and treat them equally during both training and inference, which compromises their robustness when the RGB modality becomes unreliable or unavailable.}
In this case, we propose \textbf{InfraNet}, an IR-centric quality-aware framework that regulates RGB guidance during training and supports flexible RGB--IR or IR-only deployment.
InfraNet employs an asymmetric architecture where the primary IR pathway extracts multi-scale infrared features for predictions, while the auxiliary RGB pathway provides reliability-controlled supervisory signals. The core of InfraNet is \textbf{QualGate}, a quality-aware fusion module that learns a task-oriented control signal to suppress unreliable RGB guidance and compensate IR features during cross-modal training.
Built upon InfraNet, we design two architectural variants: a lightweight IR-only architecture InfraNet-IR and an RGB--IR architecture InfraNet-RGB-IR.
Our method is evaluated through extensive experiments on four benchmark datasets (LLVIP, FLIR-Aligned, M$^3$FD, and DroneVehicle), showing strong or competitive accuracy in challenging low-light and adverse weather conditions.
Notably, InfraNet maintains high efficiency in IR-only inference, making it both accurate and computationally efficient.
\end{abstract}


\section{Introduction}
\label{sec:intro}

Object detection is a crucial task in computer vision, with widespread applications in fields such as remote sensing \cite{GuiSQT24}, video surveillance \cite{PayghodeGBID23}, and autonomous driving \cite{yang2025review}. 
While significant advancements have been made in detection algorithms, many current approaches still rely predominantly on RGB images \cite{10309914, Hussain24}. 
Nevertheless, they face performance challenges in difficult conditions, such as low-light environments, glare, and adverse weather.
This is due to the inherent dependency of RGB sensors on ambient light, which limits their effectiveness in poorly lit settings or highly fluctuating lighting conditions. 
To address these limitations, infrared (IR) sensing offers a robust alternative by capturing the thermal radiation emitted by objects, enabling consistent detection even under poor visibility or severe lighting variations \cite{KristoIP20}.
While IR imaging provides stable responses under low-light and adverse weather conditions, RGB imaging contributes rich spatial and color details under favorable illumination. 
Building upon this complementarity, multi-modal object detection has emerged as an effective paradigm that integrates RGB and IR modalities to achieve more reliable perception across diverse environments. 
Depending on the integration stage, cross-modality fusion networks are generally categorized into pixel-level fusion \cite{ijcai2020didfuse, pr2024icafusion}, feature-level fusion \cite{wavemamba,fusionmamba}, and decision-level fusion frameworks \cite{tcsvt2024evidential, tmm2024m2fnet}.

\begin{wrapfigure}{r}{0.48\textwidth} 
  \vspace{-1.0cm}
  \centering
  \includegraphics[width=0.48\textwidth]{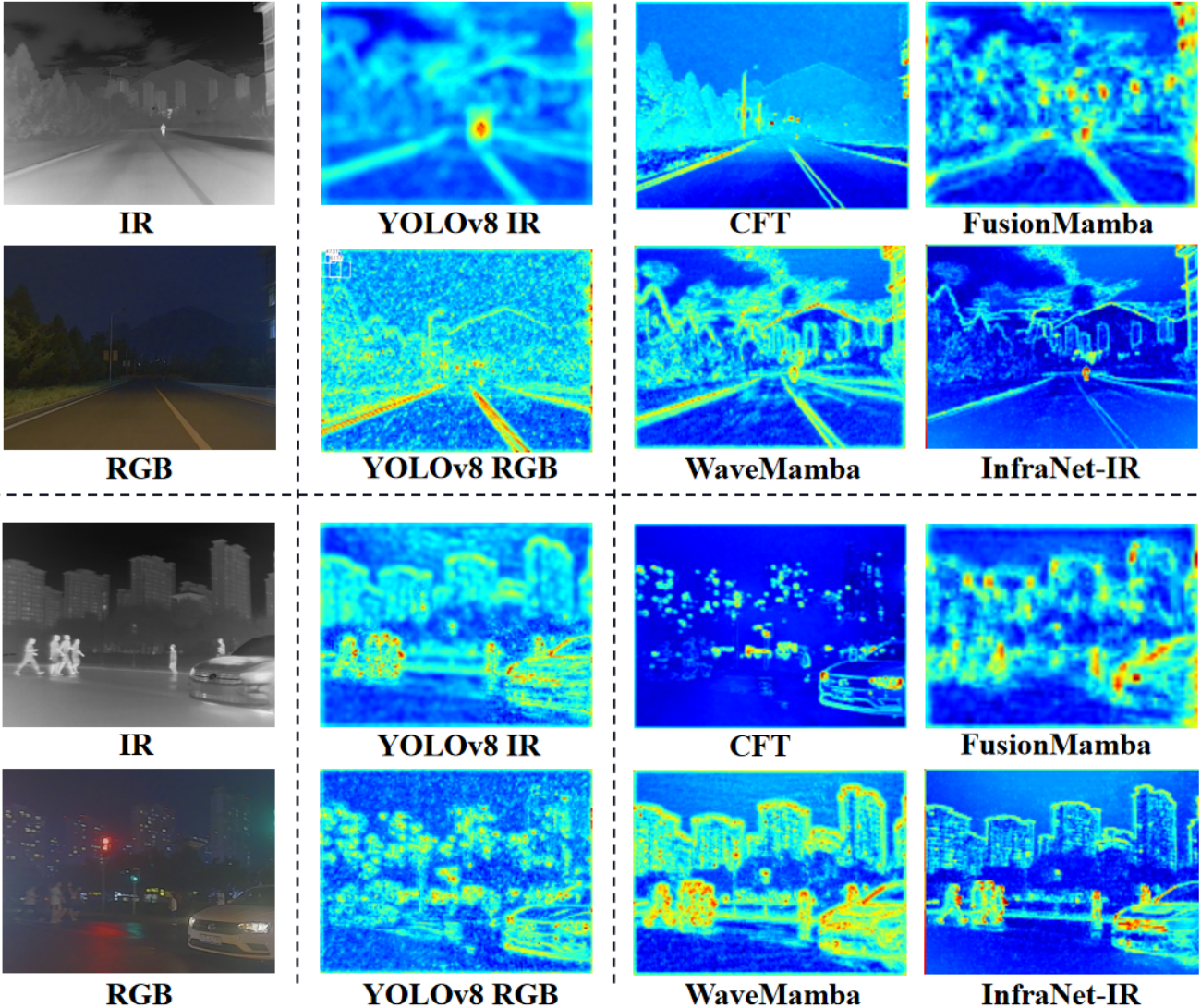}
  \caption{
    Visualization of RGB and IR feature quality under nighttime conditions on M$^3$FD dataset. While RGB features exhibit substantial noise and degradation, IR features remain robust and clear. Existing fusion methods (CFT, FusionMamba, WaveMamba) introduce additional noise compared to pure IR features, motivating our IR-centric design that regulates RGB guidance during training.
  }
  \label{fig:problem_statement}
\vspace{-0.5cm}
\end{wrapfigure}
While significant progress has been made in multi-modal object detection, most current methods treat the RGB and IR modalities equally, assuming that both provide complementary information to enhance detection performance. 
However, these approaches still overlook a crucial issue: the unreliability of the RGB modality under challenging conditions.
In environments with poor lighting or adverse weather, RGB images can become severely degraded, often resulting in entirely dark or noisy images with little discernible object structure.
When such degraded RGB features are indiscriminately fused with the more reliable IR features, they can introduce irrelevant or even harmful information, ultimately impairing the overall detection accuracy.
As illustrated in Fig.~\ref{fig:problem_statement}, we visualize IR and RGB backbone features from a YOLOv8 model trained on the M$^3$FD dataset using channel summation for feature visualization. Under nighttime conditions, the RGB features exhibit substantial noise, while the IR features remain relatively intact.
Additionally, when comparing traditional fusion methods such as CFT~\cite{qingyun2021fusing}, FusionMamba~\cite{fusionmamba}, and WaveMamba~\cite{wavemamba}, we visualize their fused backbone features after RGB-IR training and find that the fusion process increases the noise level compared to the original IR features, further undermining the reliability of the fusion outcome.
Building on these insights, \textit{we aim to prioritize the more reliable IR modality, while selectively incorporating RGB features only when they provide useful and stable information, thereby enhancing detection robustness and accuracy. }

{
In this paper, we propose \textbf{InfraNet}, {a quality-aware framework for flexible RGB-IR/IR-only object detection.}
The core of InfraNet lies in \textbf{QualGate}, a quality-aware fusion module designed to regulate cross-modal guidance.
Deployed at different scales, QualGate learns a task-oriented reliability control signal from RGB features. It suppresses unreliable RGB contributions and compensates the IR features during cross-modal training, providing more stable supervisory signals for the IR-centered representation.
{
Built upon InfraNet, we design two architectural variants: 
\textbf{(1) InfraNet-IR:} A lightweight single-branch network optimized for computational efficiency, trained with RGB auxiliary supervision but deployed with IR-only inference. 
\textbf{(2) InfraNet-RGB-IR:} A higher-capacity dual-branch structure that processes both RGB and IR inputs during both training and inference, utilizing QualGate modules for reliability-controlled cross-modal fusion.
These two architectures are implemented and trained as two independent networks.
}
Extensive experiments conducted on four benchmark datasets demonstrate that InfraNet achieves strong accuracy for both dual-modality and IR-only inference.
Notably, it maintains high efficiency in the IR-only setting and remains robust in challenging low-light and adverse weather conditions.}
{In summary, our main contributions are summarized below:
\begin{itemize}
\item {We propose \textbf{InfraNet}, a quality-aware RGB-guided framework for infrared object detection. InfraNet uses RGB as reliability-controlled auxiliary guidance during training to enhance the learning of IR-centered features, while allowing IR-only deployment.
}
\item {We design \textbf{QualGate}, a quality-aware fusion module that learns a task-oriented control signal to regulate cross-modal contributions by suppressing unreliable RGB guidance and compensating IR features to mitigate the risk of negative transfer.
}
\item {
Extensive experiments on four RGB-IR benchmarks show that InfraNet achieves competitive or superior performance across different deployment settings, including efficient IR-only inference and higher-capacity RGB-IR inference.
} 
\end{itemize}
}

\section{Related Works}
\label{sec:related}

\subsection{RGB-Infrared Multi-modal Object Detection}

RGB-infrared (IR) multi-modal detection leverages complementary characteristics: RGB provides rich texture under favorable illumination, while thermal IR maintains robust signatures in darkness and adverse weather~\cite{jia2021llvip,liu2022target,sun2022drone}.
Existing fusion paradigms can be categorized into pixel-level, feature-level, and decision-level strategies, each with distinct trade-offs.
\textbf{Pixel-level fusion}
operates in the raw image domain by merging RGB and IR inputs before feature extraction.
Representative approaches leverage adversarial learning~\cite{ijcai2020didfuse,liu2022target}, correlation-driven decomposition~\cite{cvpr2023cddfuse}, or semantic-aware integration~\cite{jas2022superfusion,if2023divfusion,pr2024icafusion} to combine complementary information.
Despite preserving fine-grained details, this paradigm introduces substantial preprocessing overhead and struggles with high-level semantic alignment, frequently resulting in inferior detection accuracy.
\textbf{Feature-level fusion}
extracts hierarchical representations via dual-stream backbones and integrates cross-modal features at intermediate network layers.
Pioneering works employ cross-modal refinement~\cite{ijcv2021sdnet} or attention mechanisms~\cite{cvpr2022rfnet,mm2022detfusion,mm2023ignet} to enhance feature complementarity.
State-of-the-art methods leverage coarse-to-fine fusion strategies~\cite{rsdet2024}, Transformer architectures~\cite{prl2024crossformer}, or state-space models with wavelet transforms~\cite{wavemamba}, achieving remarkable performance gains.
YOLO-based extensions~\cite{rs2024acdfyolo,s24134098,10555757} adapt single-stage detectors with multi-modal backbones for efficient deployment.
A critical limitation, however, is that all these approaches mandate \emph{dual-modal input at inference}, rendering them impractical when RGB sensors are unavailable or heavily corrupted.
\textbf{Decision-level fusion}
processes each modality independently through separate detectors and ensembles predictions via score aggregation or uncertainty weighting~\cite{pr2019illumination,eccv2022shanet,tcsvt2024evidential,tmm2024m2fnet}.
This paradigm offers robustness to single-modal failure and accommodates heterogeneous architectures, yet sacrifices cross-modal representation learning during feature extraction.
Consequently, decision-level fusion often underperforms feature-level methods when both modalities are reliable, while incurring doubled computational cost due to parallel detector deployment.
{
Unlike conventional fusion that treats two modalities equally, we propose a quality-aware auxiliary supervision paradigm to enhance the reliability of cross-modal fusion, supporting flexible architecture design.
Specifically, built upon InfraNet, we design two architectural variants, one for IR-only inference and the other for RGB-IR inference.
}

\subsection{Auxiliary Learning and Adaptive Fusion}

Auxiliary branch learning leverages additional network branches to enhance the main task through complementary supervision during training.
Related paradigms include learning using privileged information (LUPI)~\cite{vapnik2009lupi} and knowledge distillation~\cite{hinton2015distilling}, which transfer knowledge from auxiliary signals or pre-trained teachers.
However, these approaches assume fixed privileged information or employ fixed teacher models, limiting their flexibility when the reliability or availability of modalities changes.
Adaptive fusion mechanisms handle quality variations through illumination-aware weighting~\cite{tits2023mfpt,grsl2024essfn} or attention-based interactions~\cite{cvprw2023csaa,tiv2024mmsanet,cvpr2024textif}.
Yet most methods perform feature-level weighting without considering the deployment-oriented asymmetry between training-time guidance and test-time inputs.

Recent RGB-T/RGB-X works have also explored dynamic or reliability-aware fusion at the feature level, including scene-specific fusion and adaptive weighting mechanisms for adverse conditions and cross-modal interaction \cite{bijelic2020seeing,deevi2024rgbx,zhang2021abmdrnet,li2025crossmodalnet,xu2025mambatrans,medeiros2025mipa}. While effective, most of these methods are designed for dual-modal inference and primarily focus on reweighting or enhancing fusion itself. In contrast, InfraNet-IR follows a privileged-guidance setting, where RGB is used only during training and can be completely removed at inference. We also relate our design to reliability-learning modules such as CRLM and CAGF in RGB-T tracking/segmentation \cite{liu2023quality,zhang2021abmdrnet} and LUPI-style methods such as HalluciDet \cite{Medeiros_2024_WACV}. Compared with these approaches, our key distinction is a lightweight scalar control score $q$ that regulates \emph{RGB suppression} and \emph{IR compensation}, together with a practical IR-only export path for deployment.

In this paper, we propose a quality-aware auxiliary supervision framework that uses RGB as task-oriented guidance for IR representation learning.
Instead of treating RGB as a required test-time modality, InfraNet-IR removes the RGB branch and all fusion modules at deployment, while InfraNet-RGB-IR keeps both modalities for higher-capacity dual-modal inference.

\begin{figure*}[!t]
  \centering
  \includegraphics[width=1.0\textwidth]{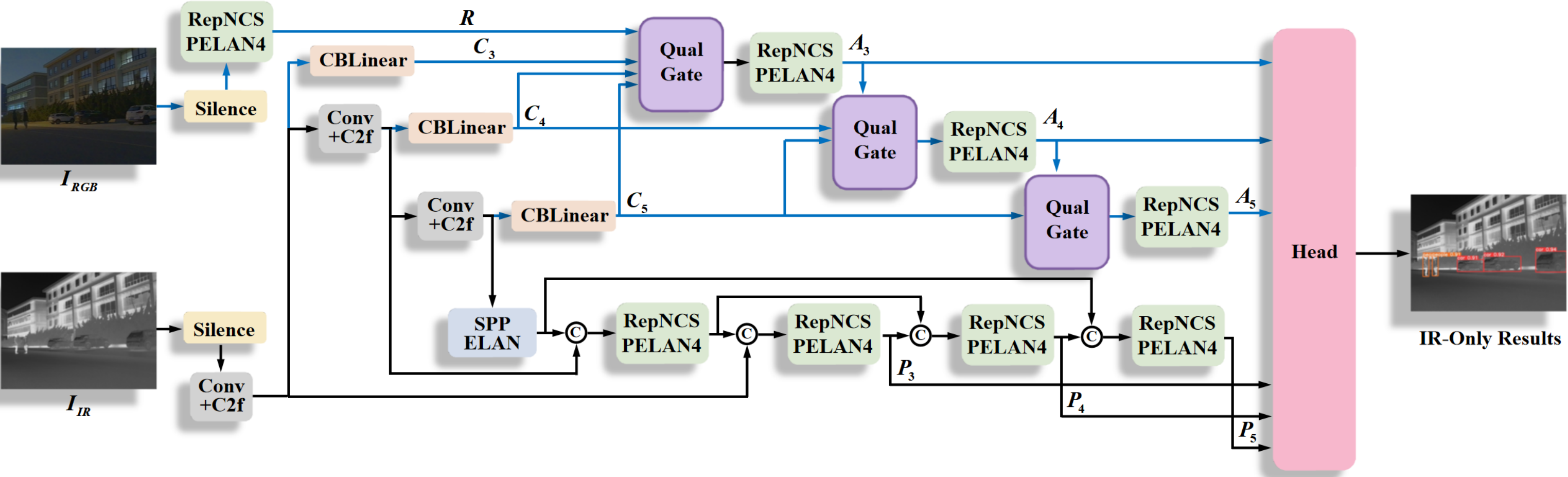}
  \vspace{-0.2cm}
  \caption{
 {
 Overview of InfraNet-IR (YOLOv9-based). During training, a main IR branch (black) and an auxiliary RGB branch (blue) interact via QualGate modules at three fusion sites (A3, A4, A5). Through quality-aware suppression/amplification, QualGate constrains the cross-branch interaction to promote IR-consistent representations while preventing degraded RGB cues from contaminating the IR stream, which justifies removing the RGB branch at test time. At inference, only the main IR branch and detection head are retained, yielding a pure IR detector with zero additional overhead.
  }
  }
  \label{fig:architecture}
\vspace{-0.6cm}
\end{figure*}

\section{Method}
\label{sec:method}
\subsection{Quality-Aware Fusion Module}
\label{subsec:fusion_modules}

\begin{figure}[!t]
  \centering
  \includegraphics[width=0.7\textwidth]{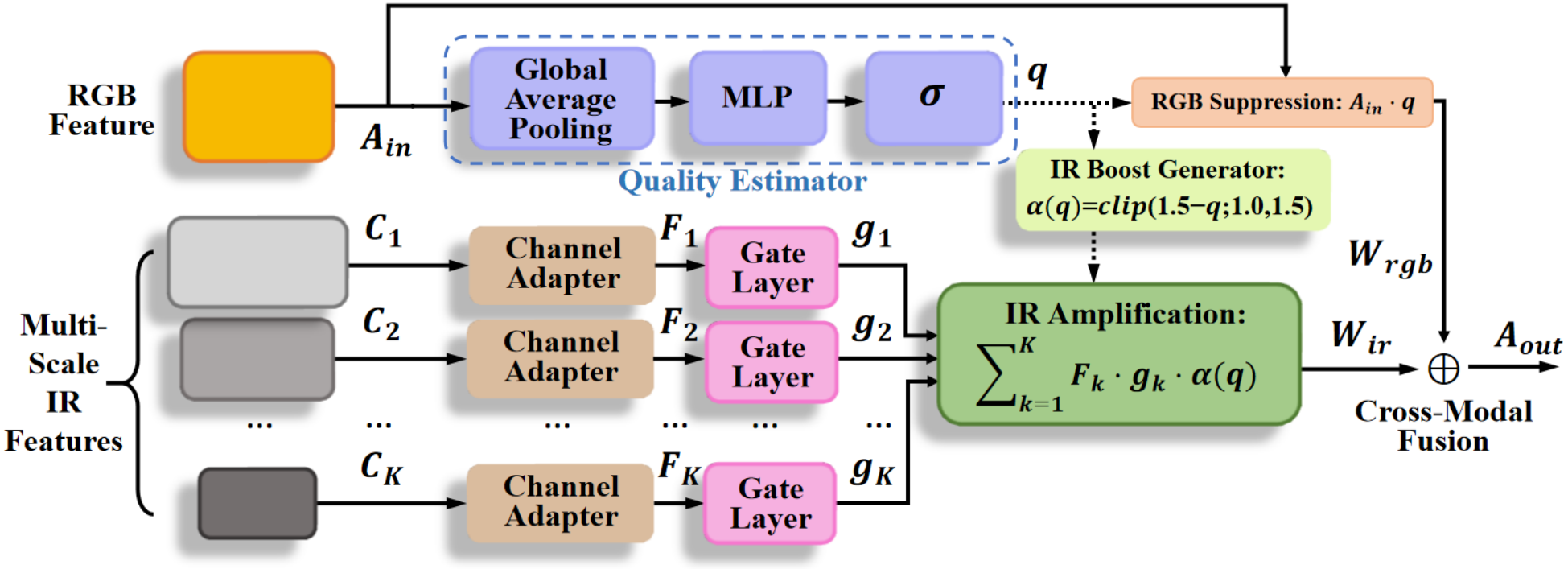}
  \caption{
  {Architecture of QualGate.}
  It takes as input $K$ scales IR features $\{C_{k}\}_{k=1}^{K}$ and the auxiliary feature $A_{in}$, and targets a quality-aware fused output $A_{out}$. For fusion, the control branch predicts a scalar score $q$ for auxiliary guidance and compensates IR features accordingly based on $\alpha(q)$.}
  \label{fig:qualgate}
\vspace{-0.3cm}
\end{figure}

RGB informativeness varies drastically across training samples: it remains rich and informative in daylight conditions but degrades significantly under low-light, fog, or motion blur scenarios.
Naive fusion methods that treat RGB as equally reliable across all conditions inevitably inject noise when RGB quality is poor, contaminating otherwise informative IR features and destabilizing training.
To address this, we propose \emph{QualGate}, a quality-aware fusion module that learns a task-oriented reliability control signal and regulates cross-modal guidance for robust representation learning.
As illustrated in Fig.~\ref{fig:qualgate}, QualGate takes $K$ scales IR feature $\{C_{k}\}_{k=1}^{K}$, and the auxiliary feature $A_{in}$ which can be RGB feature  or fused feature as input, and targets quality-aware fused output $A_{out}$.

We instantiate QualGate at each fusion site to perform \emph{dual-path reliability modulation}: 
(i) \emph{RGB suppression} to prevent negative transfer from unreliable auxiliary cues, and 
(ii) \emph{IR amplification} to compensate the IR stream when RGB guidance becomes weak. 
Given multi-scale IR features $\{C_k\}_{k=1}^{K}$ and an auxiliary feature $A_{in}$ (from the RGB branch at the current site), QualGate outputs a quality-aware fused feature $A_{out}$.

\paragraph{Reliability control score.}
We predict a scalar reliability-control score $q\in(0,1)$ from $A_{in}$:
\begin{equation}
q = \sigma\!\Big(\mathrm{MLP}_{\mathrm{qual}}\big(\mathrm{GAP}(A_{in})\big)\Big),
\label{eq:quality_score_new}
\end{equation}
where $\mathrm{MLP}_{\mathrm{qual}}$ is a lightweight two-layer bottleneck network. 
We adopt a scalar design for efficiency and optimization stability, consistent with our IR-centric deployment goal. Here, $q$ is optimized for the detection objective and is used as a control signal for RGB guidance, rather than as a hand-crafted image-quality metric.

\paragraph{RGB suppression gate.}
The auxiliary contribution is explicitly suppressed by $w_{\mathrm{rgb}}(q)$, implemented as
$w_{\mathrm{rgb}}(q)=q$, such that unreliable RGB guidance is down-weighted via $w_{\mathrm{rgb}}(q)\cdot A_{in}$.

\paragraph{IR amplification gate.}
In parallel, we apply an inverse reliability gain to the IR stream:
\begin{equation}
\alpha(q)=\mathrm{clip}\!\big(1.5-q;\ 1.0,1.5\big),
\label{eq:ir_amplifier_new}
\end{equation}
which increases IR contribution when RGB is degraded. This simple clipped form is used as a default non-suppressive IR compensation rule; we do not assume the constant 1.5 to be uniquely optimal.

\paragraph{Multi-scale IR injection and fusion.}
For each scale, we align IR features by a channel adapter and apply a content gate:
\begin{equation}
F_k=\mathrm{CA}(C_k),\qquad 
g_k=\sigma\!\Big(\mathrm{MLP}_{\mathrm{gate}}\big(\mathrm{GAP}(F_k)\big)\Big).
\end{equation}
Finally, QualGate fuses the suppressed auxiliary term with the amplified IR injection:
\begin{equation}
A_{out}=w_{\mathrm{rgb}}(q)\cdot A_{in}+\sum_{k=1}^{K}\alpha(q)\cdot g_k\cdot F_k.
\label{eq:qual_gate_fusion_new}
\end{equation}
A larger $q$ preserves more auxiliary guidance and keeps the IR gain close to nominal scale; 
a smaller $q$ reduces the auxiliary contribution and increases the IR compensation, yielding more stable IR representation shaping for IR-only inference.

\paragraph{Discussion.}
This dual-path design is empirically necessary: removing either the suppression or amplification branch leads to consistent drops (Table~\ref{tab:qualgate_dissection}), and the learned $q$ exhibits task-oriented behavior under controlled RGB degradations (Table~\ref{tab:q_degradation}). 

\subsection{InfraNet}

To avoid degraded RGB contaminating informative IR features and inducing negative transfer, we propose \textbf{InfraNet}, a quality-aware RGB guided framework based on QualGate. 
InfraNet is a plug-and-play design for common multi-scale detectors such as ResNet~\cite{he2015deep} and YOLOv5/v8/v12~\cite{jocher2022ultralytics}, by preserving the dual-branch structure and strategically placing QualGate modules at corresponding multi-scale fusion sites.
{Based on InfraNet, we design 
two deployment configurations: (1) InfraNet-IR, a lightweight IR-only architecture for efficient single-modal inference, and (2) InfraNet-RGB-IR, a higher-capacity dual-branch architecture for RGB--IR fusion. Both configurations utilize QualGate modules for quality-aware feature processing.
}

{Fig.~\ref{fig:architecture} illustrates InfraNet-IR, which built on YOLOv9 components (\ie, Conv, C2f, SPPELAN, RepNCSPELAN4)~\cite{wang2024yolov9}.} Given paired RGB images $I_{\text{RGB}}$ and infrared images $I_{\text{IR}}$, our InfraNet adopts two asymmetric pathways: a main IR branch ({black arrows}) and an auxiliary RGB branch ({{blue arrows}}). 
The IR pathway encodes multi-scale features $\{P_3,P_4,P_5\}$ for detection via successive Conv+C2f and SPPELAN stages, producing hierarchical representations.  
Meanwhile, the RGB pathway extracts base features through a lightweight stem and a RepNCSPELAN4 block.
At three designated fusion sites, the QualGate modules adaptively integrate RGB cues with the corresponding IR features, yielding $\{A_3,A_4,A_5\}$. The RGB guidance strength is regulated by the learned control score.
The fused features are then processed by RepNCSPELAN4 blocks and passed to the detection head. 
{
 For InfraNet-IR, the detection head processes both feature sets ($\{A_3,A_4,A_5\}$ and $\{P_3,P_4,P_5\}$) during training, a lightweight network processes only IR features ($\{P_3,P_4,P_5\}$) during inference, without requiring RGB input or fusion components.
}

For the design of InfraNet-RGB-IR, we retain the full dual-stream backbone–neck with RGB–IR inputs and multi-scale QualGate fusion during inference, where RGB and IR flows jointly produce and fuse the feature maps, resulting in a higher-capacity RGB–IR detector than the pruned InfraNet-IR variant. More details can be found in the supplementary.

\section{Training and Inference}
\label{sec:training_inference}


\subsection{Training}
\label{subsec:training}
Given paired inputs $(I_{\mathrm{IR}}, I_{\mathrm{RGB}})$, the InfraNet extracts a multi-scale hierarchy $\{P_3,P_4,P_5\}$ from the main branch and $\{A_3,A_4,A_5\}$ through the quality-aware fusion mechanism. Based on the detection head, standard detection losses (\ie, box regression, classification, and distribution-based localization) are employed under task-aligned assignment at each scale.
Let $\mathcal{L}^{(\mathrm{main})}_{\mathrm{det}}$ and $\mathcal{L}^{(\mathrm{aux})}_{\mathrm{det}}$ denote the main and auxiliary branch detection losses, respectively. 
For both architectural variants (\ie, InfraNet-IR and InfraNet-RGB-IR), we employ the same asymmetric weights,
\begin{equation}
\mathcal{L} = \mathcal{L}^{(\mathrm{main})}_{\mathrm{det}} \;+\; w_{\mathrm{aux}}\, \mathcal{L}^{(\mathrm{aux})}_{\mathrm{det}},
\label{eq:total_loss}
\end{equation}
with $w_{\mathrm{aux}}$ set to 0.25 by default. 
This choice ensures that optimisation is dominated by the infrared pathway, while the auxiliary branch provides reliability-adaptive guidance that enhances infrared representations.


\subsection{Inference}
\label{subsec:inference}


{
The two architectural variants, InfraNet-IR and InfraNet-RGB-IR, follow different inference settings.

\noindent\textbf{InfraNet-IR.} During training, InfraNet-IR takes paired RGB and IR inputs, where RGB serves as privileged auxiliary guidance for shaping the IR representation. During inference, only IR is used for prediction: the RGB auxiliary branch and all fusion operators are removed, and only the infrared backbone-neck hierarchy $\{P_s\}$ and the detection head are retained.


\noindent\textbf{InfraNet-RGB-IR.} InfraNet-RGB-IR adopts 
a dual-stream backbone that processes both RGB and IR inputs during training and inference. QualGate continues to regulate RGB--IR interactions at multiple scales.

}

\section{Experiments}
\label{sec:experiments}

\subsection{Experimental Setup}
\label{subsec:setup}

\paragraph{Datasets.}

We evaluate on four established RGB–IR benchmarks: \textbf{LLVIP}~\cite{jia2021llvip} (15{,}488 aligned image pairs; pedestrian detection), \textbf{FLIR-Aligned}~\cite{flir2018} (5{,}142 aligned pairs; automotive detection with three classes), \textbf{M$^3$FD}~\cite{liu2022target} (4{,}200 pairs spanning six classes under diverse conditions), and \textbf{DroneVehicle}~\cite{sun2022drone} (28{,}439 UAV-captured pairs; five-class vehicle detection).

\paragraph{Implementation.}
All experiments are conducted on a single NVIDIA RTX 4090 GPU with PyTorch.
Models are trained for 300 epochs using SGD (momentum 0.937, weight decay $5{\times}10^{-4}$) with cosine learning rate decay ($10^{-2}\rightarrow10^{-4}$) and batch size 32 under the single-GPU setting.
Dual-branch training employs $w_{\mathrm{aux}}=0.25$.
Standard YOLO augmentations (Mosaic, MixUp, HSV, horizontal flip, scaling) are applied, with geometric transforms synchronized across modalities to preserve spatial alignment. 
We evaluate the models based on InfraNet-IR and InfraNet-RGB-IR, to demonstrate the flexibility of our framework, separately.
{For simplicity, we use Ours (IR) to represent InfraNet-IR, and Ours (RGB-IR) to represent InfraNet-RGB-IR.
}
This ensures consistent experimental conditions across all main comparisons and ablation studies.


\paragraph{Metrics.}
We report standard metrics $mAP_{50}$ (IoU=0.5) and $mAP$ (IoU=0.5:0.95).
Besides, for FLIR, we report additional Precision, Recall, and F1-score; for
M$^3$FD and DroneVehicle, we report per-class $AP_{50}$.

\paragraph{Training overhead and convergence.}
QualGate is lightweight, implemented as two-layer MLPs.
For example, during training, InfraNet-IR (YOLOv8) introduces an auxiliary RGB branch (26.1M parameters) for privileged guidance, while at inference it keeps an IR-only model (28.2M) with \emph{no extra deployment cost}.
In practice, the added training overhead is acceptable---the model can be trained on a single consumer-grade GPU.
Moreover, under the standard 300-epoch protocol, our method converges stably and typically reaches a stable performance regime around 130 epochs.

\subsection{Comparison with State-of-the-Art Methods}
\label{subsec:sota}

\paragraph{LLVIP.}

\begin{wraptable}{r}{0.5\columnwidth}
  \vspace{-1.2cm}
  \centering
  \small
  \caption{Comparison results with SOTA methods on LLVIP dataset. The best results are highlighted in \textcolor{red}{red}. The second and third best results are highlighted in \textcolor{green}{green} and \textcolor{blue}{blue}, respectively.}
  \scalebox{0.62}{
  \begin{tabular}{ccc|cc}
    \toprule
    Methods & Mode & Backbone & $mAP_{50}$ & $mAP$ \\
    \midrule
    (ACM MM'23) TIRDet~\cite{wang2023tirdet} & IR & CSPD53 & 96.3 & 64.2 \\
    (WACV'24) HalluciDet~\cite{Medeiros_2024_WACV} & IR &Faster RCNN & 88.3 & - \\
    \midrule
    (CVPRW'23) CSAA~\cite{cvprw2023csaa} & RGB+IR & ResNet50 & 94.3 & 54.2 \\
    (2024) RSDet~\cite{rsdet2024} & RGB+IR & ResNet50 & 95.8 & 61.3 \\
    (TIV'24) MMSANet~\cite{tiv2024mmsanet} & RGB+IR & ResNet50 & 96.4 & 62.5 \\
    (ICCV'25) WaveMamba~\cite{wavemamba} & RGB+IR & ResNet50 & 97.3 & 65.0 \\
    \textbf{Ours (RGB-IR)} & RGB+IR & ResNet50 & \textcolor{green}{98.1} & 69.2 \\
    \textbf{Ours (IR)} & IR & ResNet50 & 97.7 & 68.4 \\
    \midrule
    (CVPR'24) Text-IF~\cite{cvpr2024textif} & RGB+IR & Transformer & 94.1 & 60.2 \\
    \midrule
    (IF'23) DIVFusion~\cite{if2023divfusion} & RGB+IR & YOLOv5 & 89.8 & 52.0 \\
    (PR'24) ICAFusion~\cite{pr2024icafusion} & RGB+IR & YOLOv5 & 95.2 & 60.1 \\
    (ICCV'25) WaveMamba~\cite{wavemamba} & RGB+IR & YOLOv5 & 97.6 & 65.2 \\
    \textbf{Ours (RGB-IR)} & RGB+IR & YOLOv5 & \textcolor{red}{98.3} & \textcolor{blue}{70.2} \\
    \textbf{Ours (IR)} & IR & YOLOv5 & 97.3 & 68.1 \\
    \midrule
    (ICHMS'24) MDMDet~\cite{10555757} & RGB+IR & YOLOv7 & 96.5 & 61.5 \\
    \midrule
    YOLOv8l-IR~\cite{10533619} & IR & YOLOv8 & 95.2 & 62.1 \\
    YOLOv8l-RGB~\cite{10533619} & RGB & YOLOv8 & 91.9 & 54.0 \\
    (RS'24) ACDF-YOLO~\cite{rs2024acdfyolo} & RGB+IR & YOLOv8 & 96.5 & 61.3 \\
    (ICHMS'24) MSFF~\cite{10555757} & RGB+IR & YOLOv8 & 96.8 & 61.9 \\
    (Sensors'24) FAWDet~\cite{s24134098} & RGB+IR & YOLOv8 & 97.1 & 62.1 \\
    (ICCV'25) WaveMamba~\cite{wavemamba} & RGB+IR & YOLOv8 & \textcolor{red}{98.3} & 66.0 \\
    \textbf{Ours (RGB-IR)} & RGB+IR & YOLOv8 & \textcolor{green}{98.1} & \textcolor{green}{70.5} \\
    \textbf{Ours (IR)} & IR & YOLOv8 & 97.3 & 68.8 \\
    \midrule
    \textbf{Ours (RGB-IR)} & RGB+IR & YOLOv12 & \textcolor{blue}{98.0} & \textcolor{red}{70.6} \\
    \textbf{Ours (IR)} & IR & YOLOv12 & 97.2 & 67.6 \\
    \bottomrule
  \end{tabular}}
  \label{tab:example5}
  \vspace{-0.8cm}
\end{wraptable}
Table~\ref{tab:example5} presents pedestrian detection results on the LLVIP dataset. Our method achieves strong performance across different backbone architectures. Specifically, our RGB--IR variants attain \textbf{70.5\% mAP} (YOLOv8) and \textbf{70.6\% mAP} (YOLOv12), improving over the strong recent method WaveMamba (66.0\%). Notably, our IR-only deployment also demonstrates strong performance, with ResNet50-IR achieving 68.4\% mAP while requiring no additional computational cost at inference, thereby validating the effectiveness of our auxiliary supervision approach.

\begin{table*}[!t]
  \centering
  \small
    \caption{Comparison results with SOTA methods on FLIR-Aligned dataset. The best results are highlighted in \textcolor{red}{red}, the second and third best results are highlighted in \textcolor{green}{green} and \textcolor{blue}{blue}, respectively. For Precision/Recall/F1/$mAP_{50}$/\,$mAP$, larger is better; for Parameters and Inference time, smaller is better.}
  \scalebox{0.7}{
  \begin{tabular}{ccc|ccc|cc|cc}
    \toprule
    Methods & Mode & Backbone & Precision & Recall & F1 & $mAP_{50}$ & $mAP$ & Parameters & Inference time (ms) \\
    \midrule
    (ACM MM'23) TIRDet~\cite{wang2023tirdet} & IR & CSPD53 & - & - & - & 81.4 & 44.3 & - & - \\
    \midrule
    (TITS'23) MFPT~\cite{tits2023mfpt} & RGB+IR & ResNet50 & 79.3 & 72.9 & 76.0 & 80.0 & 41.9 & 200.0M & 80.0 \\
    (ICCV'25) WaveMamba~\cite{wavemamba} & RGB+IR & ResNet50 & 84.2 & 77.9 & 80.9 & 86.5 & 47.9 & 193.2M & 53.2 \\
    \textbf{Ours (RGB-IR)} & RGB+IR & ResNet50 & 83.5 & 80.4 & 81.9 & 88.6 & 54.3 & 73.8M & 14.4 \\
    \textbf{Ours (IR)} & IR & ResNet50 & 84.9 & 78.8 & 81.7 & 87.3 & 53 & 42.1M & \textcolor{red}{9.4} \\
    \midrule
    (PRL'24) CrossFormer~\cite{prl2024crossformer} & RGB+IR & YOLOv5 & 78.1 & 72.8 & 75.4 & 79.3 & 42.1 & 340.0M & 80.0 \\
    (ICCV'25) WaveMamba~\cite{wavemamba} & RGB+IR & YOLOv5 & 83.9 & 79.7 & 81.7 & 85.8 & 47.5 & 45.6M & 44.1 \\
    \textbf{Ours (RGB-IR)} & RGB+IR & YOLOv5 & \textcolor{green}{85.1} & \textcolor{red}{81.8} & \textcolor{red}{83.3} & \textcolor{red}{89.5} & \textcolor{blue}{55.1} & 52.2M & 13.2 \\
    \textbf{Ours (IR)} & IR & YOLOv5 & 84.9 & 80.4 & 82.6 & 88.1 & 53.8 & \textcolor{green}{26.0M} & \textcolor{green}{11.6} \\
    \midrule
    YOLOv8-IR~\cite{10533619} & IR & YOLOv8 & 75.1 & 65.3 & 69.9 & 72.9 & 38.3 & 43.7M & 22.0 \\
    YOLOv8-RGB~\cite{10533619} & RGB & YOLOv8 & 71.6 & 62.2 & 66.6 & 66.3 & 28.2 & 43.7M & 22.0 \\
    (GRSL'24) ESSFN~\cite{grsl2024essfn} & RGB+IR & YOLOv8 & 81.4 & 73.5 & 77.2 & 80.8 & 42.3 & 80.2M & 47.0 \\
    (ICCV'25) WaveMamba~\cite{wavemamba} & RGB+IR & YOLOv8 & 84.2 & \textcolor{blue}{80.9} & 82.5 & 88.4 & 48.1 & 69.1M & 40.0 \\
    \textbf{Ours (RGB-IR)} & RGB+IR & YOLOv8 & \textcolor{red}{85.8} & 79.9 & \textcolor{blue}{82.8} & \textcolor{green}{89.4} & \textcolor{green}{55.2} & 54.3M & 13.4 \\
    \textbf{Ours (IR)} & IR & YOLOv8 & 84.6 & \textcolor{blue}{80.9} & 82.7 & 88.5 & 54.0 & \textcolor{blue}{28.2M} & 12.5 \\
    \midrule
    \textbf{Ours (RGB-IR)} & RGB+IR & YOLOv12 & 84.3 & \textcolor{green}{81.7} & \textcolor{green}{83.0} & \textcolor{blue}{89.3} & \textcolor{red}{55.8} & 56.8M & 13.2 \\
    \textbf{Ours (IR)} & IR & YOLOv12 & \textcolor{blue}{85.0} & 79.5 & 82.2 & 87.7 & 53.2 & \textcolor{red}{24.7M} & \textcolor{blue}{12.0} \\
    \bottomrule
  \end{tabular}}
  \label{tab:example6}
  \vspace{-0pt}
\end{table*}
\begin{table*}[!t]
\centering
\small
\caption{Comparison results with SOTA methods on DroneVehicle dataset. The best results are highlighted in \textcolor{red}{red}. The second and third best results are highlighted in \textcolor{green}{green} and \textcolor{blue}{blue}, respectively. }
\scalebox{0.7}{
\begin{tabular}{ccc|cc|ccccc}
\toprule
Methods & Mode & Backbone & $mAP_{50}$ & $mAP$ & Car & Bus & Truck & Freight-car & Van \\
\midrule
(ECCV'22) TSFADet~\cite{10.1007/978-3-031-20077-9_30} & RGB+IR & ResNet50 & 73.1 & 44.1 & 89.9 & 89.8 & 67.9 & 63.7 & 54.0 \\
(GRSL'24) GLFNet~\cite{10476333} & RGB+IR & ResNet50 & 71.4 & 54.8 & 90.3 & 88.0 & 72.7 & 53.6 & 52.6 \\
(TGRS'24) LF-MDet~\cite{10643097} & RGB+IR & ResNet50 & 71.8 & 51.3 & -- & -- & -- & -- & -- \\
(TGRS'24) $C^2$Former-$S^2$ANet~\cite{10472947} & RGB+IR & ResNet50 & 74.2 & 47.3 & 90.2 & 89.8 & 68.3 & 64.4 & 58.5 \\
(2024) DMM~\cite{zhou2024dmmdisparityguidedmultispectralmamba} & RGB+IR & ResNet50 & 77.2 & 55.8 & 90.4 & 88.7 & 77.8 & 63.0 & 66.0 \\
(JST'24) FFODNet~\cite{10461034} & RGB+IR & ResNet50 & 76.3 & 56.9 & -- & -- & -- & -- & -- \\
(ICCV'25) WaveMamba~\cite{wavemamba} & RGB+IR & ResNet50 & 79.3 & 59.9 & 94.6 & 90.2 & 79.6 & 68.0 & 64.1 \\
\textbf{Ours (RGB-IR)} & RGB+IR & ResNet50 & 83.9 & \textcolor{blue}{64.1} & \textcolor{green}{98.0} & \textcolor{red}{97.3} & 80.5 & \textcolor{blue}{78.0} & 65.7 \\
\textbf{Ours (IR)} & IR & ResNet50 & 83.5 & 63.0 & \textcolor{blue}{97.8} & \textcolor{blue}{97.1} & 79.3 & 75.0 & \textcolor{blue}{68.3} \\
\midrule
(YAC'24) CAFN-IA~\cite{10598791} & RGB+IR & YOLOv5 & 69.3 & 56.1 & 89.1 & 90.8 & 62.0 & 57.3 & 47.1 \\
(RSL'24) multimodal DINO~\cite{article} & RGB+IR & YOLOv5 & 72.5 & 50.3 & 89.5 & 88.8 & 75.4 & 54.3 & 54.3 \\
(MTA'23) SLBAF-Net~\cite{10.1007/s11042-023-15333-w} & RGB+IR & YOLOv5 & 76.8 & 49.5 & 90.2 & 89.9 & 76.0 & 68.2 & 59.9 \\
(CVPR'24) CSOM-ODAF~\cite{10655285} & RGB+IR & YOLOv5 & 77.1 & 50.1 & 90.1 & 89.8 & 75.6 & 68.2 & 61.8 \\
(ICCV'25) WaveMamba~\cite{wavemamba} & RGB+IR & YOLOv5 & 79.5 & 60.2 & 94.8 & 90.1 & 80.1 & 68.2 & 64.3 \\
\textbf{Ours (RGB-IR)} & RGB+IR & YOLOv5  & 84.1 & 63.8 & \textcolor{red}{98.1} & \textcolor{green}{97.2} & \textcolor{green}{82.1} & 76.2 & 67.1 \\
\textbf{Ours (IR)} & IR & YOLOv5  & 83.6 & 63.1 & \textcolor{green}{98.0} & 96.9 & 79.6 & 75.5 & 68.2 \\
\midrule
(Sensors'23) Dual-YOLO~\cite{s23062934} & RGB+IR & YOLOv7 & 71.9 & 55.2 & 95.9 & 91.6 & 69.7 & 55.9 & 46.6 \\
\midrule
YOLOv8l-IR~\cite{10533619} & IR & YOLOv8 & 71.9 & 49.1 & 93.4 & 91.9 & 69.3 & 53.7 & 51.1 \\
YOLOv8l-RGB~\cite{10533619} & RGB & YOLOv8 & 70.2 & 48.6 & 92.5 & 91.7 & 68.8 & 47.8 & 50.2 \\
(AEORS'24) YOLOFIV~\cite{10643643} & RGB+IR & YOLOv8 & 64.7 & 53.1 & 95.9 & 91.6 & 64.2 & 34.6 & 37.3 \\
(TGRS'24) CRSIOD~\cite{10440361} & RGB+IR & YOLOv8 & 73.2 & 50.8 & 95.6 & 92.2 & 71.7 & 50.5 & 55.8 \\
(IJAEOG'24) M2FNet~\cite{JIANG2024103918} & RGB+IR & YOLOv8 & 76.8 & 50.4 & 95.8 & 93.1 & 75.9 & 59.8 & 59.4 \\
(Sensors'24) IV-YOLO~\cite{s24196181} & RGB+IR & YOLOv8 & 74.6 & 56.8 & 97.2 & 94.3 & 65.4 & 63.1 & 53.0 \\
(RS'24) DAAB-FFPN~\cite{rs16203904} & RGB+IR & YOLOv8 & 75.2 & 56.3 & -- & -- & -- & -- & -- \\
(ICCV'25) WaveMamba~\cite{wavemamba} & RGB+IR & YOLOv8 & 79.8 & 60.5 & 95.0 & 90.6 & 80.4 & 68.5 & 64.5 \\
\textbf{Ours (RGB-IR)} & RGB+IR & YOLOv8  & \textcolor{red}{85.0} & \textcolor{red}{64.9} & \textcolor{red}{98.1} & \textcolor{red}{97.3} & \textcolor{red}{82.6} & \textcolor{red}{79.9} & 67.3 \\
\textbf{Ours (IR)} & IR & YOLOv8  & 84.0 & 63.4 & \textcolor{green}{98.0} & 97.0 & 80.0 & 76.9 & \textcolor{green}{68.4} \\
\midrule
\textbf{Ours (RGB-IR)} & RGB+IR & YOLOv12 & \textcolor{green}{84.8} & \textcolor{green}{64.7} & \textcolor{green}{98.0} & \textcolor{red}{97.3} & \textcolor{red}{82.6} & \textcolor{green}{78.3} & 67.6 \\
\textbf{Ours (IR)} & IR & YOLOv12 & \textcolor{blue}{84.7} & 63.9 & \textcolor{red}{98.1} & \textcolor{red}{97.3} & \textcolor{blue}{81.3} & 77.7 & \textcolor{red}{69.3} \\
\bottomrule
\end{tabular}}
\label{tab:example4}
\vspace{-10pt}
\end{table*}

\paragraph{FLIR-Aligned.}
Table~\ref{tab:example6} presents automotive thermal detection results on the FLIR-Aligned dataset. Our method achieves strong performance across multiple metrics: YOLOv5-RGB-IR attains the highest recall (81.8\%), F1-score (83.3\%), and mAP$_{50}$ (89.5\%), while YOLOv12-RGB-IR achieves the best mAP (55.8\%). Notably, our IR-only variants demonstrate superior efficiency: YOLOv12-IR requires only 24.7M parameters and ResNet50-IR achieves 9.4ms inference time, substantially outperforming competing fusion methods while maintaining competitive accuracy. This validates our auxiliary supervision paradigm that achieves strong performance without deployment overhead.

\begin{figure*}[!t]
  \centering
  \includegraphics[width=0.8\textwidth]{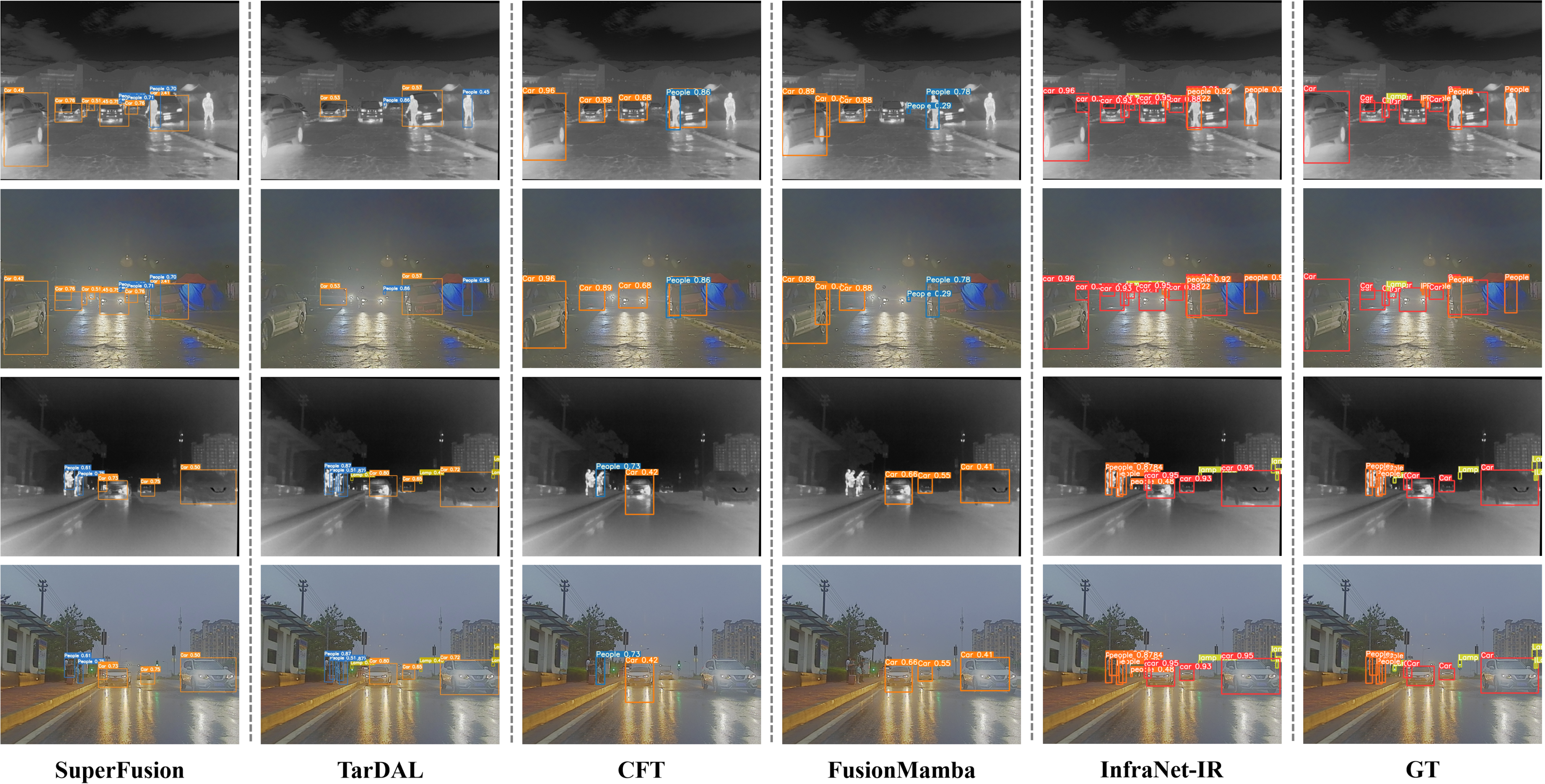}
  \caption{
   Visualization of detection results of several cross-modality object detection methods on M$^3$FD dataset. InfraNet-IR produces tighter bounding boxes, fewer false positives, and more consistent detections across diverse illumination conditions. Unlike prior methods (SuperFusion, TarDAL, CFT, FusionMamba), which often suffer from RGB degradation and background leakage, InfraNet-IR maintains robust localization by relying on IR-dominant features enhanced through quality-aware supervision. GT annotations are provided in the last column.
  }
  \label{fig:detection_result}
\vspace{-0.3cm}
\end{figure*}

\begin{wraptable}{r}{0.5\textwidth}  
  \vspace{-1cm}  
  \centering
  \small
  \caption{Comparison results on $M^3$FD subsets in AP50. The best results are highlighted in \textcolor{red}{red}, the second best results are highlighted in \textcolor{blue}{blue}, respectively.}
  \scalebox{0.65}{
    \begin{tabular}{lc|cccc}
    \hline
    Method & Backbone & Day & Overcast & Night & Challenge \\
    \hline
    Infrared~\cite{liu2022target} & YOLOv5 & 58.9 & 80.3 & 79.5 & 70.9 \\
    Visible~\cite{liu2022target} & YOLOv5 & 59.1 & 82.4 & 78.7 & 75.9 \\
    DenseFuse~\cite{li2019densefuse} & Dense-CNN & 60.8 & 75.9 & 80.6 & 83.7 \\
    FusionGAN~\cite{ma2019fusiongan} & CNN & 60.1 & 81.6 & 79.8 & 66.7 \\
    RFN~\cite{cvpr2022rfnet} & RFN & 59.2 & 79.6 & 80.3 & 82.7 \\
    GANMcC~\cite{ma2020ganmcc} & CNN & 60.3 & 79.6 & 81.1 & 82.7 \\
    DDcGAN~\cite{li2019ddcgan} & CNN & 59.4 & 78.0 & 77.1 & 68.9 \\
    MFEIF~\cite{zhou2020mfeif} & CNN & 60.7 & 77.0 & 81.2 & 68.3 \\
    U2Fusion~\cite{xu2020u2fusion} & VGG-16 & 60.4 & 79.3 & 78.3 & 83.6 \\
    TarDAL~\cite{liu2022target} & YOLOv5 & 61.3 & 82.3 & 81.6 & 84.6 \\
    \textbf{Ours (RGB-IR)} & YOLOv5 & \textcolor{blue}{{87.6}} & 86.8 & 96.6 & \textcolor{red}{{99.5}} \\
    \textbf{Ours (IR)} & YOLOv5 & 82.7 & 81.2 & 92.3 & \textcolor{red}{{99.5}} \\
    \textbf{Ours (RGB-IR)} & YOLOv8 & \textcolor{red}{{87.7}} & \textcolor{blue}{87.4} & \textcolor{red}{{96.8}} & \textcolor{red}{{99.5}} \\
    \textbf{Ours (IR)} & YOLOv8 & 83.5 & 81.4 & 92.1 & \textcolor{red}{{99.5}} \\
    \textbf{Ours (RGB-IR)} & YOLOv12 & \textcolor{blue}{{87.6}} & \textcolor{red}{{88.7}} & \textcolor{blue}{{96.7}} & \textcolor{red}{{99.5}} \\
    \textbf{Ours (IR)} & YOLOv12 & 81.2 & 80.8 & 92.3 & \textcolor{blue}{99.4} \\
    \hline
    \end{tabular}
  }
  \label{tab:m3fd_4}
\vspace{-10pt}
\end{wraptable}

\paragraph{DroneVehicle.}
Table~\ref{tab:example4} presents UAV-based 5-class vehicle detection results on the DroneVehicle dataset.
Our YOLOv12-IR achieves the best results among the compared methods with \textbf{84.7\% $mAP_{50}$} and \textbf{63.9\% mAP}, surpassing WaveMamba (79.8\%/60.5\%) by substantial margins of \textbf{+4.9\%/ +3.4\%} (+6.1\%/+5.6\% relative)—the largest absolute gains across all datasets.
Per-class analysis shows particularly strong performance on challenging small-object categories: \textbf{Car} 98.1\% $AP_{50}$ (rank 1st), \textbf{Bus} 97.3\% (1st), \textbf{Freight-car} 77.7\% (+9.2\% over WaveMamba), \textbf{Van} 69.3\% (+4.8\%).
These substantial small-object improvements stem from our hierarchical multi-scale IR tip aggregation: at the finest detection scale A3, fusing P3/P4/P5 features provides robust representations resilient to RGB degradation from atmospheric scattering at high altitudes—a critical advantage for aerial
\noindent imaging where RGB visibility is severely compromised by weather, haze, and altitude-induced degradation.

\begin{figure}[!t]
  \centering
  \includegraphics[width=0.6\textwidth]{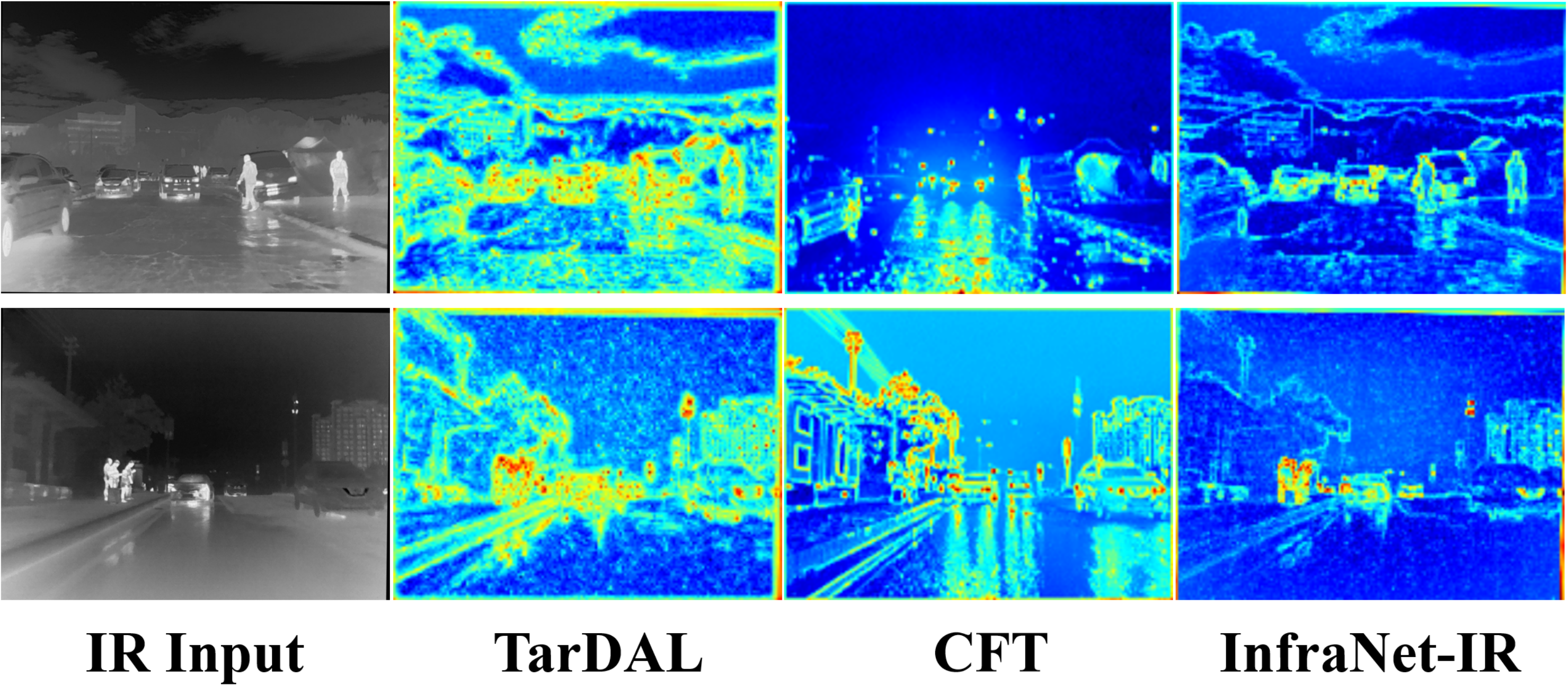}
  \caption{
  Feature-level visualization on M$^3$FD nighttime examples. InfraNet-IR produces more target-aligned responses with less background activation. This visualization is used to diagnose the effect of RGB-guided training under degraded RGB conditions; the deployed InfraNet-IR model still performs IR-only inference.
  }
  \label{fig:feature}
 \vspace{-8pt}
\end{figure}

\paragraph{M$^3$FD.}
We first evaluate our model on the M$^3$FD benchmark, and complete quantitative results are provided in the Appendix.
To further characterize performance under different illumination conditions, we also report results on the commonly used Day, Overcast, Night, and Challenge subsets of M$^3$FD.
As shown in Table~\ref{tab:m3fd_4}, InfraNet obtains strong results across these subsets, especially in night scenes where RGB visibility is severely compromised.
Since many recent RGB--IR detection methods do not report results under exactly the same subset protocol, Table~\ref{tab:m3fd_4} is used as a diagnostic comparison rather than an exhaustive current-SOTA ranking.
The IR-only variants also maintain strong night performance, indicating that RGB-guided training can improve the IR representation without requiring RGB at deployment.
Fig.~\ref{fig:detection_result} shows that our model achieves tighter localization and fewer false positives than representative fusion baselines. To further inspect the feature behavior under degraded RGB conditions, we visualize intermediate features in Fig.~\ref{fig:feature}. These visualizations are used as training-time feature-level diagnostics; the deployed InfraNet-IR model still uses only IR input.

\subsection{Ablation Studies}
\label{subsec:ablation}

We conduct comprehensive ablation studies to validate our design choices and dissect the contribution of each component.
Unless otherwise specified, all ablations use YOLOv8 with consistent training protocols (300 epochs, SGD optimizer, standard augmentation).

\paragraph{QualGate Design Dissection.}

Table~\ref{tab:qualgate_dissection} isolates the contribution of QualGate from both the \emph{module} and \emph{mechanism} perspectives. 
Replacing QualGate with HalluciDet-style pseudo-RGB guidance or existing adaptive weighting modules (CAGF/CRLM) leads to inferior results, indicating that generic reliability-aware fusion is insufficient for our IR-centric setting. 
More importantly, removing either branch of the proposed modulation degrades performance: disabling IR amplification (\(\alpha(q)\equiv 1\)) causes a clear drop, and disabling RGB suppression (\(w_{\mathrm{rgb}}\equiv 1\)) also hurts accuracy. 
These results support our central design choice that a learned scalar reliability score \(q\) should jointly \emph{suppress unreliable RGB guidance} and \emph{compensate the IR stream}, which is critical for robust IR representation learning and IR-only deployment.

\vspace{-0.6em}
\begin{table}[ht]
\centering
\setlength{\tabcolsep}{4pt}
\renewcommand{\arraystretch}{1.05}
\small
\caption{QualGate design dissection on LLVIP using YOLOv8 (InfraNet-IR setting). 
We compare QualGate with representative alternatives (HalluciDet-style pseudo-RGB guidance, CAGF, and CRLM), and further ablate its two key components: RGB suppression and IR amplification. 
The full QualGate consistently achieves the best performance, validating that the gain comes from the proposed \emph{suppress+amplify} reliability modulation rather than generic feature reweighting.}
\vspace{-0.6em}
\scalebox{0.8}{
\begin{tabular}{lcc}
\toprule
Setting/Backbone & $mAP_{50}$ & $mAP_{50\text{--}95}$ \\
\midrule
\midrule
\multicolumn{3}{l}{
\textbf{A. QualGate ablation (LLVIP, YOLOv8-IR)}}\\
\midrule
HalluciDet (Faster RCNN)~\cite{Medeiros_2024_WACV} & 88.3 & - \\
Ours training with pseudo-RGB by HalluciDet & 95.2 & 64.9 \\
Ours replacing QualGate with  CAGF~\cite{zhang2021abmdrnet} & 95.7 & 66.8\\
Ours replacing QualGate with  CRLM~\cite{liu2023quality}& 96.9 & 66.7 \\
Ours with Full QualGate  & \textbf{97.3} & \textbf{68.8} \\
~~w/o IR amp ($\alpha(q)\equiv 1$) & 97.1 & 67.2 \\
~~w/o RGB supp ($w_{\mathrm{rgb}}\equiv 1$) & 97.1 & 68.4 \\
~~Naive ($q\equiv 1$) & 96.9 & 67.7 \\
\midrule
\bottomrule
\end{tabular}}
\label{tab:qualgate_dissection}
\vspace{-0.6cm}
\end{table}

\paragraph{Controlled RGB Degradation and \(q\) Validation.}

Table~\ref{tab:q_degradation} analyzes the behavior of the learned scalar control score \(q\) under controlled RGB perturbations, without using any extra quality annotations.
Under fog corruption, \(q\) decreases mildly as degradation becomes stronger, while our method remains more stable than the naive variant, especially at severe degradation levels.
For brightness perturbation, \(q\) varies more mildly and is not intended to serve as a handcrafted image-quality metric.
These results support our interpretation of \(q\) as a task-oriented reliability control signal for limiting harmful RGB guidance, rather than as an explicit estimator of perceptual RGB quality.
Additional analyses on reliability granularity, fusion-site placement, and IR amplification range are provided in the supplementary material.

\begin{table}[t]
\centering
\setlength{\tabcolsep}{4pt}
\renewcommand{\arraystretch}{1.05}
\small
\caption{Controlled RGB degradation analysis on LLVIP with fixed IR input. 
We progressively degrade only the RGB modality (Fog / Brightness) and report the predicted quality score \(q\) together with detection performance (AP50/AP). 
``Naive'' disables reliability control by forcing \(q\equiv 1\). The results show that learned reliability control improves robustness under degraded RGB conditions, without requiring \(q\) to be interpreted as a perceptual image-quality score.}
\vspace{-0.6em}
\scalebox{0.75}{
\begin{tabular}{c ccc c ccc}
\toprule
& \multicolumn{3}{c}{\emph{Fog}} & & \multicolumn{3}{c}{\emph{Brightness}} \\
\cmidrule{2-4} \cmidrule{6-8}
Level & $q$ & Ours & Naive & & $q$ & Ours & Naive \\
\midrule
0.2 & 0.441 & 97.0/67.6 & 95.8/66.9 & & 0.459 & 97.5/67.8 & 96.7/66.7 \\
0.4 & 0.439 & 97.0/67.7 & 95.6/66.1 & & 0.465 & 97.5/67.6 & 96.8/67.0 \\
0.6 & 0.437 & 97.1/67.7 & 95.3/65.9 & & 0.473 & 97.3/67.1 & 96.2/65.3 \\
0.8 & 0.435 & 97.1/67.7 & 94.4/64.5 & & 0.481 & 96.9/66.9 & 95.2/65.4 \\
1.0 & 0.434 & 97.1/67.8 & 92.6/61.8 & & 0.484 & 96.5/66.4 & 93.6/63.9 \\
\bottomrule
\end{tabular}}
\label{tab:q_degradation}
\end{table}

\paragraph{Deployment Flexibility Analysis.}

Table~\ref{tab:deployment_and_loss} (a) compares three deployment configurations on M3FD: a conventional IR baseline and our two InfraNet-based detectors (InfraNet-IR and InfraNet-RGB-IR). InfraNet-IR improves the baseline by +1.4 AP50 and +1.8 AP with the same parameters and inference time, showing that auxiliary RGB supervision strengthens IR feature learning without extra deployment cost. When RGB is available, InfraNet-RGB-IR yields larger gains (+5.1 AP50, +6.8 AP), confirming the benefit of quality-aware RGB-IR fusion.

\begin{table}[!t]
\centering
\footnotesize
\setlength{\tabcolsep}{3pt}
\caption{(a) Deployment flexibility analysis on M3FD using YOLOv8. InfraNet-IR achieves zero-overhead IR-only deployment with superior performance, while InfraNet-RGB-IR offers optional dual-modal fusion. (b) Ablation study on $w_{\mathrm{aux}}$ using YOLOv8 on LLVIP.}
\label{tab:deployment_and_loss}
\vspace{-0.2cm}
\begin{minipage}[t]{0.55\columnwidth}
  \centering
  \textbf{(a)}\\[3pt]
  \scalebox{0.75}{
  \begin{tabular}{@{}lccccc@{}}
    \toprule
    Method & \multicolumn{2}{c}{M$^3$FD Performance} & \multicolumn{3}{c}{Computational Efficiency} \\
    \cmidrule(lr){2-3} \cmidrule(l){4-6}
     & AP$_{50}$ & AP & Params (M) & Speed (ms) \\
    \midrule
    IR baseline & 84.8 & 57.2 & 28.2 & 12.5 \\
    \midrule
    \textbf{Ours (IR)} & {86.2} & {59.0} & {28.2} & {12.5} \\
    \textbf{Ours (RGB-IR)} & {89.9} & {64.0} & 54.3 & 13.4 \\
    \bottomrule
  \end{tabular}}
\end{minipage}%
\hfill
\begin{minipage}[t]{0.43\columnwidth}
  \centering
  \textbf{(b)}\\[3pt]
  \scalebox{0.75}{
  \begin{tabular}{@{}cccc@{}}
    \toprule
    Backbone & $w_{\mathrm{aux}}$ & AP$_{50}$ & AP \\
    \midrule
    YOLOv8 & 0.0 & 97.0 & 68.0 \\
    YOLOv8 & 0.1 & 97.2 & 68.2 \\
    YOLOv8 & \textbf{0.25} & \textbf{97.3} & \textbf{68.8} \\
    YOLOv8 & 0.5 & 97.1 & 68.3 \\
    YOLOv8 & 1.0 & 97.1 & 68.1 \\
    \bottomrule
  \end{tabular}}
\end{minipage}
\vspace{-0.5cm}
\end{table}

\paragraph{Auxiliary Branch Loss Weight.}
Table~\ref{tab:deployment_and_loss} (b) evaluates the auxiliary branch loss weight $w_{\mathrm{aux}}$ on LLVIP. $w_{\mathrm{aux}}=0.25$ achieves optimal performance (68.8\% mAP), indicating a balanced trade-off between the use of RGB guidance and the prevention of interference.
When $w_{\mathrm{aux}}$ is too small (0.0--0.1) or too large (0.5--1.0), performance degrades (68.0--68.3\% mAP), validating our asymmetric loss weighting strategy.

\section{Conclusion}

In this paper, we introduce InfraNet, a modality-asymmetric quality-aware framework for RGB--IR/IR-only object detection.
Based on InfraNet, we design two separate architectures, the IR-only detector (InfraNet-IR) and the dual-branch RGB--IR detector (InfraNet-RGB-IR).
In both architectures, we keep IR as the primary pathway and use RGB as reliability-controlled guidance when available.
This cross-modal guidance is regulated by \textbf{QualGate}, which learns a task-oriented control signal to suppress unreliable RGB guidance and compensate IR features during training.
Through extensive experiments on four benchmark datasets, we show that InfraNet achieves strong performance under different deployment settings. Notably, InfraNet-IR maintains high inference efficiency by removing the RGB branch and fusion modules at deployment, while InfraNet-RGB-IR provides a higher-capacity option when both modalities are available.



\section*{Acknowledgements}
The work was supported by the National Key Research and Development Program of China (No.2023YFC3306401). This research was also supported by Zhejiang Provincial Natural Science Foundation of China under Grant No. LD24F020007, Beijing Natural Science Foundation L244043.

%
%
\bibliographystyle{splncs04}
\bibliography{main}
\end{document}